\documentclass[12pt]{elsarticle}
\usepackage{amssymb}
\usepackage{amsmath}
\usepackage{subfig}
\usepackage{xcolor}
\usepackage{multirow}
\usepackage{hyperref}
\usepackage{verbatim}
\usepackage{adjustbox}
\usepackage{booktabs}
\usepackage[edges]{forest}
\usepackage{acronym}
\usepackage{pgfplots}
\pgfplotsset{compat=1.18}
\usepackage{graphicx}

\makeatletter
\def\ps@pprintTitle{ }
\makeatother

\newcommand{\dbname}{BirdsEye-RU}
\newcommand{\dblink}{\url{https://www.kaggle.com/datasets/mdahanafarifkhan/birdseye-ru}}

\begin{document}

\begin{frontmatter}
\title{{\dbname}: A Dataset For Detecting Faces from Overhead Images}

\author[label1,equal]{Md. Ahanaf Arif Khan (ahanaf019@gmail.com)}
\author[label1,equal]{Ariful Islam (ariful111866@gmail.com)}
\author[label1]{Sangeeta Biswas\corref{cor1} (sangeeta.cse@ru.ac.bd)}
\author[label1]{Md. Iqbal Aziz Khan (iqbal\_aziz\_khan@ru.ac.bd)}
\author[label1]{Subrata Pramanik (sprmnk@ru.ac.bd)}
\author[label1]{Sanjoy Kumar Chakravarty (sanjoy.cse@ru.ac.bd)}
\author[label1]{Bimal Kumar Pramanik (bkp@ru.ac.bd)}

\affiliation[label1]{organization={Department of Computer Science and Engineering},
                addressline={Faculty of Engineering, University of Rajshahi},
                city={Rajshahi-6205},
                country={Bangladesh}}

\fntext[equal]{These authors contributed equally to this work.}
\cortext[cor1]{Corresponding author: Sangeeta Biswas (sangeeta.cse@ru.ac.bd).}

\begin{abstract}
Detecting faces in overhead images remains a significant challenge due to extreme scale variations and environmental clutter. To address this, we created the {\dbname} dataset, a comprehensive collection of 2,978 images containing over eight thousand annotated faces. This dataset is specifically designed to capture small and distant faces across diverse environments, containing both drone images and smartphone-captured images from high altitude. We present a detailed description of the {\dbname} dataset in this paper. We made our dataset freely available to the public, and it can be accessed at {\dblink}.
\end{abstract}

\end{frontmatter}

\section{Introduction}
\label{introduction}

Bird’s-eye digital images refer to overhead images captured from a high vantage point, resembling the perspective of a bird in flight. Such images can be acquired using drones, smartphones, surveillance cameras, or other imaging devices. Face detection in bird’s-eye imagery has become increasingly important due to its growing use in public safety, event monitoring, and crowd management applications. However, faces in overhead images are often very small, blurry, and partially occluded, which makes reliable face detection a challenging task for existing models.

Only a limited number of datasets are available for face detection from overhead imagery, including DroneFace~\cite{droneface}, DroneSURF~\cite{DroneSURF}, and Drone LAMS~\cite{luo2021dronelamsdronebasedface}. While these datasets have made valuable contributions to the field, they also exhibit certain limitations, such as reliance on simulated aerial images, restricted or broken access, and inactive contact channels.

To address these limitations, we introduce the {\dbname} dataset and make it freely available to the public. The dataset consists of overhead images with annotated faces captured in a wide range of environments, including city streets, rivers, rooftops, and parks. It features challenging visual conditions, such as small, partially visible, and occluded faces. To further enhance diversity in image quality and sensor characteristics, the dataset includes both drone-captured and smartphone-captured images.

\section{Dataset Description}
\begin{figure}
    \centering
    \includegraphics[width=\textwidth]{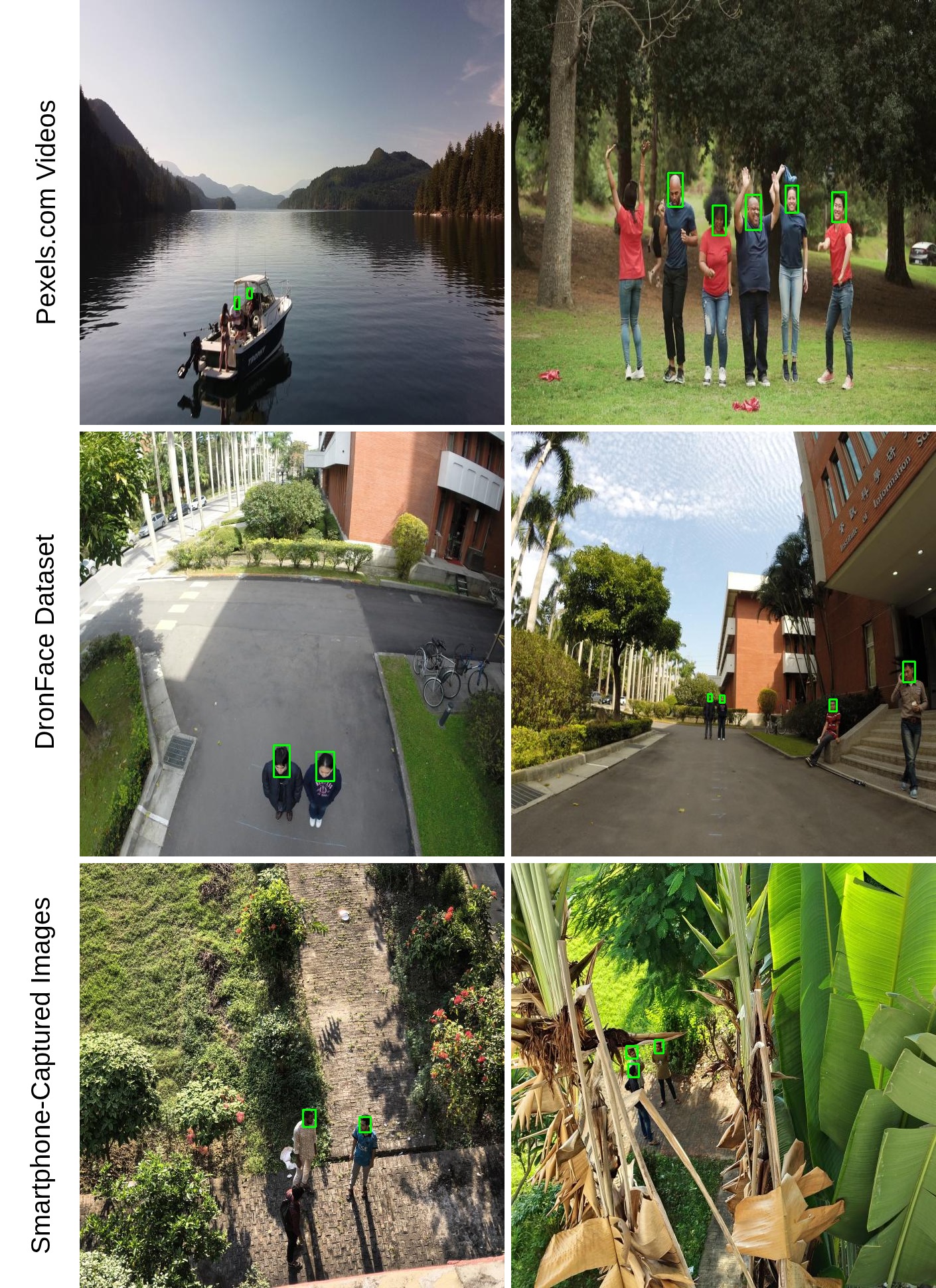}
    \caption{Representative sample images from the {\dbname} dataset with face bounding-box annotations. The rows correspond to the three sources: Pexels videos, DroneFace dataset and smartphone-captured images, respectively.}
    \label{fig:samples}
\end{figure}

The {\dbname} dataset comprises a total of 2,978 images, annotated with a bounding box for every visible face. The dataset comprises images captured by both drones and smartphones. The dataset was created using three distinct sources: videos collected from Pexels.com, images obtained from the DroneFace dataset~\cite{droneface}, and images captured using smartphone cameras. Examples with face annotations from each data source are shown in Figure~\ref{fig:samples}.

\subsection{Pexels.com Videos}
We acquired 72 drone videos of varying lengths featuring human subjects from Pexels.com, a royalty-free stock footage platform. These videos have resolutions ranging from $1280 \times 720$ to $3840 \times 2160$, with frame rates between 24 and 60 fps. From these videos, we extracted frames at a rate of one frame per second. After that, we manually curated the extracted frames to remove redundant frames and frames that did not contain any faces. After this curation process, a total of 1,601 images containing human faces were retained.

\subsection{DroneFace Dataset Images}

The DroneFace~\cite{droneface} dataset contains a total of 2,057 images including 620 raw images, 1,364 frontal face images, and 73 portrait images. It does not contain actual drone images. It simulates varying flight altitudes using a stationary GoPro camera mounted on a variable-length stick. The images were captured in a controlled setting, utilizing a total of 11 subjects and stationary poses and neutral expressions. We integrated 619 raw images with a resolution of $3680 \times 2760$ pixels and 170$^\circ$ field of view from this dataset. These raw images were captured from 1.5, 3, 4, and 5 meters high and 2 to 17 meters away.

\subsection{Our Smartphone-Captured Images}

We captured a total of 758 images using three different smartphones (a Vivo Y36, a Motorola Edge 50 Fusion, and a Motorola Edge 50 Neo). The images captured had a fixed aspect ratio of 4:3. These images were captured from various heights within the University of Rajshahi (RU) campus in uncontrolled outdoor environments and complex backgrounds. A total of 11 male subjects participated in the data collection, appearing in varying group sizes.

\section{Preprocessing and Annotation}

After acquisition, we resized all images to $640 \times 640$ pixels to align with the standard input dimensions of the YOLO architecture. Subsequently, each face was manually annotated with a bounding box using the Roboflow platform. The resulting bounding boxes follow the format \{x, y, w, h\}, where (x, y) represents the normalized center coordinates, and (w, h) indicates the normalized width and height. We annotated a total of 6,915 face instances across videos sourced from Pexels.com and images captured using smartphones. For the DroneFace dataset, we leveraged the bounding box annotations provided in~\cite{droneface_dataset} rather than re-annotating the data. All annotations were consolidated and exported in the standard YOLO format. The composition of the {\dbname} dataset is detailed in Table~\ref{tab:db_faces}.

\begin{table}[!h]
    \centering
    \caption{Distribution of images and face annotations in the {\dbname} dataset, across the three data sources. Here, \# Images: Number of Images, \# Faces: Number of Annotated Faces, and Faces/Image: Average Number of Faces Per Image.}
    \begin{tabular}{|l|r|r|r|}
    \hline
        \textbf{Data Source} & \textbf{\# Images} &\textbf{\# Faces}  &\textbf{Faces/Image}\\\hline
        Pexels.com Videos & 1,601 & 4,414  &2.76\\ 
        DroneFace Dataset & 619 & 1,533 &2.48\\
        Smartphone Images & 758 & 2,501  &3.30\\\hline
        \textbf{Total} & 2,978 & 8,448  &2.84\\\hline
    \end{tabular}
    \label{tab:db_faces}
\end{table}

\section{Dataset Split}

The {\dbname} dataset was partitioned into training, validation, and testing subsets to support model development and evaluation. The training set consists of 1,048 images extracted from 42 distinct videos from Pexels.com Videos, all 619 raw images from the DroneFace Dataset, and 334 images from Smartphone Images. The validation set includes 271 images extracted from 14 distinct videos from Pexels.com Videos and 194 images from Smartphone Images. The test set comprises 282 images extracted from 16 distinct videos from Pexels.com Videos and 230 images from Smartphone Images. The distribution of images across the three splits and data sources is summarized in Table~\ref{tab:datasets}.

\begin{table}[h]
\centering
\caption{Distribution of images across the training, validation, and testing partitions of the {\dbname} dataset.}

\resizebox{\textwidth}{!}{
\begin{tabular}{|l|r|r|r|r|}
\hline
 \multirow{2}{*}{\textbf{Partition}}& \multicolumn{3}{c|}{\textbf{Data Source}}& \multirow{2}{*}{\textbf{Total}}\\\cline{2-4}
& Pexels.com Videos& DroneFace Dataset& Smartphone Images&  \\
\hline
 Training& 1,048 & 619& 334& 2,001\\

 Validatin& 271& 0 & 194& 465\\
 Testing& 282 & 0& 230 & 512\\\hline
 \textbf{Total}& 1,601& 619& 758&2,978\\
 \hline
\end{tabular}}
\label{tab:datasets}
\end{table}

\section{Face Scale Distribution}
\begin{figure}[h]
    \centering
    \includegraphics[width=\textwidth]{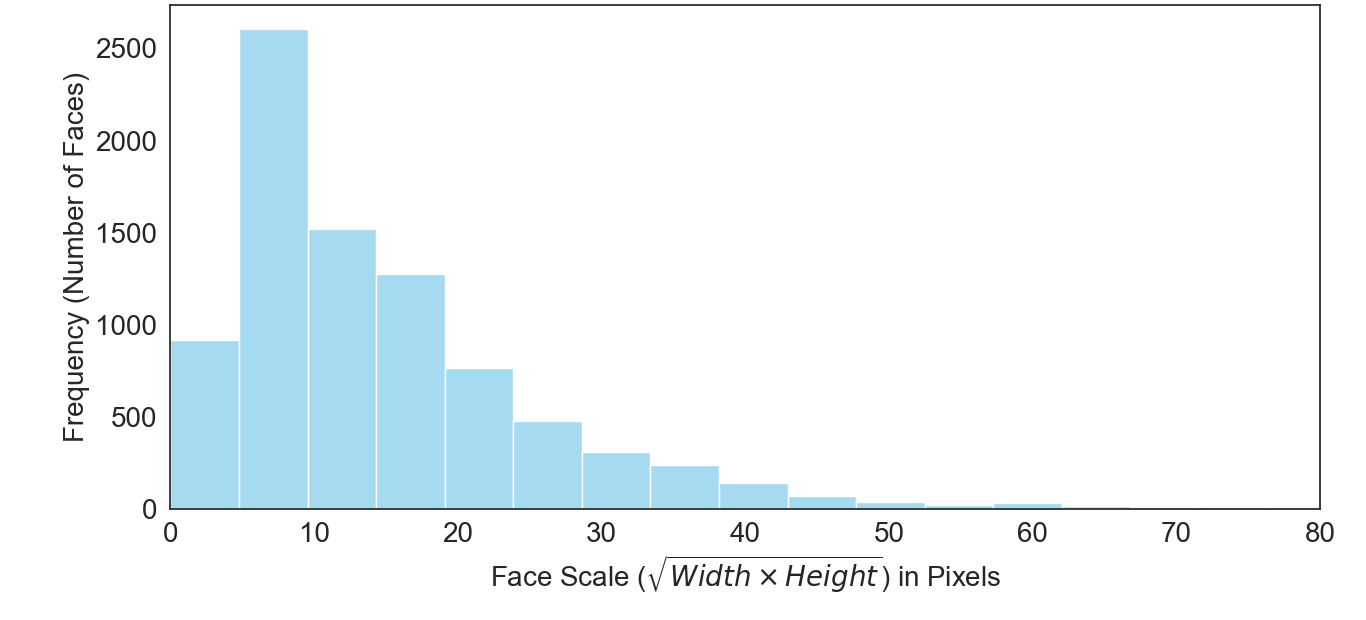}
    \caption{Distribution of face scales within the dataset. The high density of samples clustered around 10 pixels highlights the predominance of small faces.}
    \label{fig:face_size_distribution}
\end{figure}

Figure~\ref{fig:face_size_distribution} shows the face scale distribution in our dataset. The distribution shows a dominance of small-scale faces, with the peak frequency occurring at a scale of approximately 10 pixels. Since the majority of faces are smaller than 40 pixels, this dataset introduces a high level of difficulty, specifically targeting the problem of detecting low-resolution faces in high-altitude overhead images.

\section{Directory Structure}
\begin{figure}[h]
\centering
\begin{forest}
  for tree={
    font=\ttfamily,
    grow'=0,
    folder,
    s sep=-2pt,
    l sep=20pt,
  }
   [{\dbname}
        [train
            [images]
            [labels]
        ]
        [val
            [images]
            [labels]
        ]
        [test
            [images]
            [labels]
        ]
        [dataset.yaml]
        [db\_info.csv]
    ]
\end{forest}
\caption{Directory structure of the proposed dataset.}
\label{fig:folder_structure}
\end{figure}
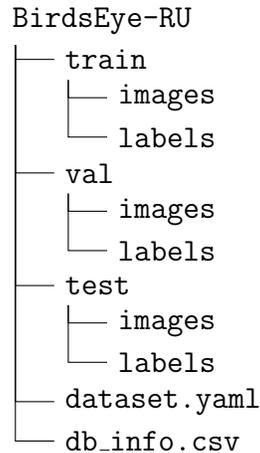

The organization of the dataset is illustrated in Figure~\ref{fig:folder_structure}. The root directory, \textit{{\dbname}}, is organized into three distinct subdirectories: \texttt{train}, \texttt{val}, and \texttt{test}. Each subdirectory maintains an identical internal structure, comprising two folders for images and labels, respectively.

In addition to the data splits, the root directory includes a configuration file, \texttt{dataset.yaml}, formatted according to the standard YOLO specification, and a metadata file, \texttt{db\_info.csv}, which provides supplementary information such as image paths, data source identifiers, and split assignments (train, validation, or test).

\section*{Ethical Considerations and Data Integrity}
\label{Ethical Considerations and Data Integrity}
Data collection was conducted in accordance with ethical guidelines and institutional policies. Participation was voluntary, and informed consent was obtained from all subjects. No personally identifiable information was retained, and the data is intended solely for academic research.

\section*{Dataset Availability}
The dataset is publicly accessible for research purposes via Kaggle at the following link: \dblink.

\newpage
\bibliographystyle{elsarticle-num} 
\bibliography{References/references}
\end{document}